\documentclass[runningheads]{llncs}

\usepackage{eccv}

\usepackage{eccvabbrv}

\usepackage{graphicx}
\usepackage{booktabs}

\usepackage[accsupp]{axessibility}  % Improves PDF readability for those with disabilities.

\usepackage[table]{xcolor} 
\usepackage{amsmath}
\usepackage{amssymb}

\usepackage{adjustbox}
\usepackage{tabularx}
\usepackage{subcaption} 

\usepackage[ruled,vlined]{algorithm2e}

\usepackage{booktabs}    % For professional table formatting
\usepackage{multirow}    % For multi-row cells
\usepackage{arydshln}    % For dashed lines
\usepackage{makecell}    % For better cell formatting
\usepackage{rotating}    % For rotated text
\usepackage{xcolor}      % For colored text
\usepackage{caption}

\newcommand{\ccol}{\cellcolor{blue!7}}

\usepackage[hidelinks]{hyperref} 
\usepackage{orcidlink}

\begin{document}

\title{Test-Time Hallucination Control in Large Vision-Language Models}

\titlerunning{TTH}

\author{Mehran Tamjidi\inst{1*\ }\orcidlink{0009-0002-4755-9613} \and
 Hamidreza Dastmalchi\inst{2*\ }\orcidlink{0000-0003-2818-4182} \and
Ali Cheraghian\inst{3}\orcidlink{0000-0002-3324-7849}
 \and
Mohammadreza Alimoradijazi\inst{4}\orcidlink{0000-0003-1099-4611}
 \and
Aijun An\inst{2}\orcidlink{0000-0003-1765-5751}
 \and
Hossein Rahmani\inst{5}\orcidlink{0000-0003-1920-0371}
}

\authorrunning{M.~Tamjidi et al.}

\institute{\textsuperscript{1 }University of Technology Sydney, \textsuperscript{2 }York University, Canada, \\ 
 \textsuperscript{3 }Australian National University, \textsuperscript{4 }The University of New South Wales, \\ \textsuperscript{5 }Lancaster University
\\
\email{mehran.tamjidi@uts.edu.au, [hrd, aan]@yorku.ca, reza.moradi@unsw.edu.au, ali.cheraghian@anu.edu.au, h.rahmani@lancaster.ac.uk}}

\maketitle
\def\thefootnote{*}\footnotetext{Equal contribution.}
\begin{abstract}

Object Hallucination in large vision-language models (LVLMs), where models generate non-factual content about input images, remains a critical barrier to their reliability in real-world applications. Existing mitigation strategies can be categorized into training-based and training-free methods. Training-based methods often achieve strong performance but are costly, requiring extensive computational resources, large-scale data, and time-consuming fine-tuning. Training-free approaches are particularly appealing due to their efficiency. However, existing training-free methods either require multiple decoding rounds, which adds computational overhead, or modify internal states in a model-specific way that risks degrading pretrained knowledge.
We propose \textbf{T}est-\textbf{T}ime \textbf{H}allucina-tion Mitigation (\textbf{TTH}) method, a novel training-free method that addresses both limitations. TTH introduces a token-validator module, implemented as a zero-shot Multi-Modal Classifier (MMC), to generate auxiliary logits grounded in the input image. These logits are fused with the original LVLM outputs at the token level for object tokens selected from a candidate pool. An entropy-based weighting scheme is then applied to enable robust and accurate predictions.
Extensive experiments across multiple LVLM families and diverse benchmarks demonstrate that TTH consistently improves accuracy and robustness, underscoring its generalizability and practical effectiveness. Code is released
at \url{https://github.com/Mehran-TAM/TTH}


\end{abstract}
    
\section{Introduction}

Recent advances in Large Vision-Language Models (LVLMs) \cite{liu2023improved, liu2024visual, ye2023mplug, zhu2023minigpt4, dai2023instructblip, chen2023shikra} have demonstrated remarkable capabilities in interpreting and reasoning over visual content through natural language. 
These models help users as visual assistants in tasks ranging from image captioning \cite{rotstein2024fusecap, lu2025benchmarking} to visual question answering \cite{shao2023prompting, lee2024visualquestionansweringinstruction, wang2024weaklysupervisedgaussiancontrastive}, enabling more effective human–machine collaboration.
However, these successes are shadowed by the undesirable object hallucination (OH), where LVLMs generate objects that do not accurately reflect the visual scene \cite{jiang2025devils, ha2026first, leng2024mitigating, min2026fine, hoang2025pas}. This phenomenon represents a critical failure mode, undermining model reliability and limiting its adoption in real-world applications. To address these challenges, researchers have recently proposed diverse approaches to mitigate hallucination \cite{liu2024survey, yang2025nullu, an2025mitigating, wang2026same, yin2026dynamic}, which can be broadly divided into two categories.

\begin{figure}[t!]
	\centering
	\captionsetup{skip=6pt} % Reduce space between image and caption
	\includegraphics[width=1
    \linewidth]{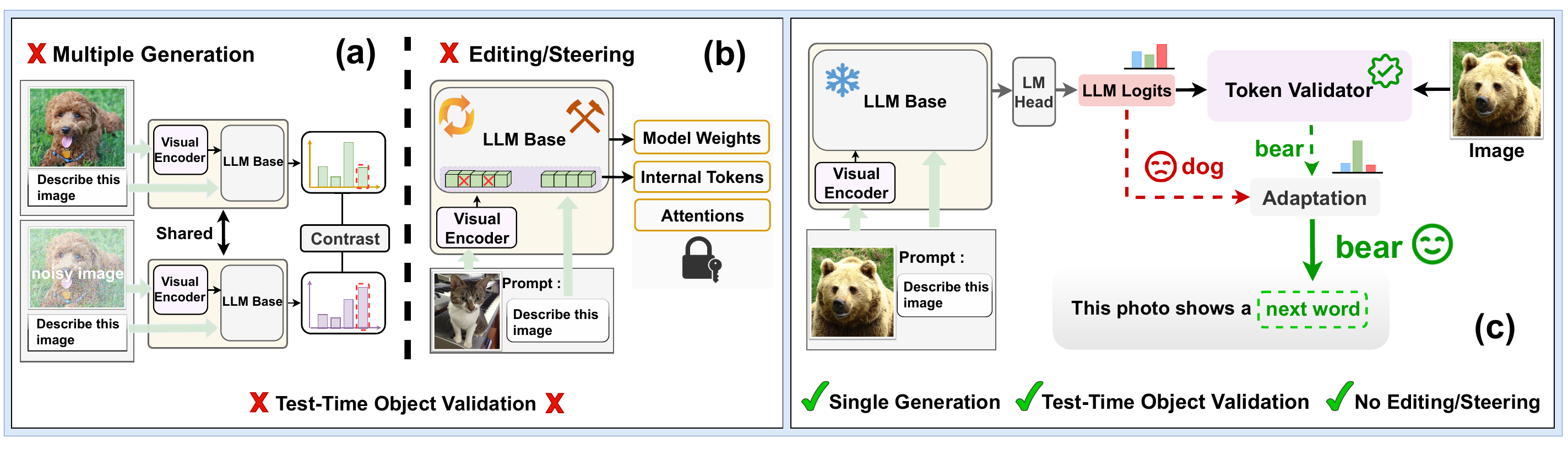}
    \caption{\textbf{(a)}
 A major category of training-free hallucination mitigation strategies relies on multiple generations during decoding, which is computationally inefficient.
\textbf{(b)} Another category modifies internal layers (editing) or hidden states (steering) before or during inference, which is often complex and may disrupt the model’s knowledge and reasoning.
\textbf{(c)} In contrast, our proposed method, TTH, overcomes both limitations by requiring only a single generation and operating without altering model weights or hidden states, making it both efficient and stable. }

	\label{fig:intro}
\end{figure}

The first category of hallucination mitigation approaches relies on extensive fine-tuning~\cite{chen2024halc, park2025convis, zhu2025ibd, leng2024mitigating, deng2024seeing, wada2025zina}, which requires carefully curated datasets and substantial computational resources. In contrast, training-free methods~\cite{an2025mitigating, jiang2025devils, chen2026residual, liu2024paying} are efficient and flexible, making them more suitable for real-world applications. By avoiding costly fine-tuning, they allow rapid adaptation across domains, driving their growing prominence in recent research.

Training-free approaches can be broadly investigated across two areas. The first group leverages model weights and internal states by weight editing \cite{yang2025nullu}, token-level manipulation \cite{zhuang2025vasparse, wang2025mint, fa2026one}, steering \cite{zhang2026prefill, dang2026locate}, and attention analysis \cite{an2025mitigating, jiang2025devils, wu2024controlmllm, liu2024paying, zhong2026adaiat}. Although sometimes effective, these methods require modifying the internal states of LVLMs, a technically complex and model-specific process that risks disrupting knowledge and reasoning abilities acquired through large-scale pretraining. In contrast, the second line of work mitigates hallucination during the decoding process \cite{chen2024halc, park2025convis, zhu2025ibd, leng2024mitigating}. These methods typically rely on contrastive or auxiliary decoding strategies, where additional generations or alternative decoding paths are used to suppress spurious outputs. While effective at reducing hallucination, such approaches introduce significant computational overhead, increasing latency and limiting their practicality in real-world applications.

In this work, we propose Test-Time Hallucination Mitigation (TTH), a light-weight and training-free strategy for reducing object hallucinations in LVLMs. TTH can be seamlessly applied to existing models without retraining, making it broadly applicable in real-world settings. Unlike existing approaches that rely on generating additional tokens—requiring multiple forward passes—or intervening in the model’s internal states, TTH operates directly on the decoding process. It leverages zero-shot multi-modal classifiers (MMCs), such as CLIP~\cite{radford2021learning} model, to validate candidate tokens during generation. Operating at the logit level, TTH selectively refines token choices with validation feedback, effectively reducing uncertainty and suppressing spurious objects before they are committed to the output sequence. This lightweight design preserves efficiency while enhancing robustness against hallucinations. As illustrated in Figure~\ref{fig:intro}, when a hallucinated object is generated (e.g., “dogs”), TTH leverages feedback from a token validator, such as CLIP, to revise the prediction. This feedback is used to adapt and refine the logits for the next token, thereby aligning the generated description more closely with the actual visual content of the image.

 Prior studies indicate that OH most often arises from low-confidence object predictions, where the model’s logit distribution is high in entropy \cite{jiang2024interpreting}. To address this, at each decoding step we construct a candidate pool consisting of object tokens (e.g., under top-k sampling) and evaluate these candidates with CLIP model. The LVLM and CLIP logits are then fused with weights determined by the inverse of the LVLM’s confidence: confident object predictions rely more on LVLM logits, while less confident ones place greater weight on CLIP. This adaptive design intensifies intervention where hallucinations are most likely, thereby suppressing spurious object predictions while preserving efficiency and accuracy in confident cases. TTH achieves these improvements with minimal overhead, sidestepping the computational inefficiencies of prior decoding methods~\cite{chen2024halc, park2025convis, zhu2025ibd, leng2024mitigating, deng2024seeing}.

\noindent Our contributions are summarized as follows:  
\begin{itemize}

    \item We introduce Test-Time Hallucination Mitigation (TTH), a decoding strategy that reduces object hallucinations in LVLMs without additional training, architectural changes, or reliance on internal model states.

    \item TTH selectively targets object tokens, identified using a WordNet-based  set \cite{miller1995wordnet}, and fuses LVLM logits with feedback from a zero-shot multi-modal classifier (e.g., CLIP). An entropy-based aggregation mechanism adaptively balances these signals, applying stronger correction to low-confidence predictions, which are more susceptible to hallucination.

    \item Extensive experiments on multiple benchmarks show that TTH consistently mitigates hallucinations in various LVLMs, including LLaVA \cite{liu2023improved, liu2024visual}, mPLUG-2 \cite{ye2024mplug}, and MiniGPT-4 \cite{zhu2023minigpt4}, while maintaining efficiency and generalizability, surpassing previous baselines.

\end{itemize}

\label{sec:intro}

\vspace{-0.2cm}

\section{Related Work}

\noindent \textbf{Large Vision-Language Models (LVLMs)} have emerged as a cornerstone of multi-modal AI by jointly advancing visual understanding and linguistic reasoning through instruction fine-tuning on diverse tasks such as image captioning \cite{rotstein2024fusecap, lu2025benchmarking} and visual question answering \cite{shao2023prompting, lee2024visualquestionansweringinstruction, wang2024weaklysupervisedgaussiancontrastive}. 
LLaVA \cite{liu2023improved, liu2024visual} stands out among these models by connecting the vision encoder \cite{radford2021learning} and the Large-Language Model (LLM) \cite{touvron2023llama2} through a linear projection that serves as an image-text alignment module. Other paradigms, including MiniGPT-4 \cite{zhu2023minigpt4}, InstructBLIP \cite{dai2023instructblip}, and mPLUG-Owl1 \cite{ye2023mplug}, utilize  Q-Former \cite{li2023blip} to unify modalities while compressing the sequence length of visual tokens. mPLUG-Owl2 \cite{ye2024mplug} further achieves effective image–text integration via adaptive cross-modal collaboration.
Beyond these paradigms, recent progress in architecture design, optimization strategies, and the use of larger-scale data has led to LVLMs such as Shikra \cite{chen2023shikra}, Qwen-VL \cite{bai2023qwen, bai2023qwenllm}, and SVIT \cite{zhao2023svitscalingvisualinstruction}, which demonstrate strong generative and reasoning abilities.
Despite their remarkable success, LVLMs still encounter significant limitations commonly stemming from reliance on co-occurring objects and biases during image-text alignment, such as object hallucination.

\vspace{0.2cm}

\noindent \textbf{Object Hallucination (OH) in LVLMs} typically refers to cases where the model generates objects that are not actually present in the input image \cite{leng2024mitigating, jiang2025devils, yang2025nullu, wu2024logical, hou2026ves, wang2026after}. 
This adverse issue raises security concerns and undermines the reliability of LVLM generations in critical applications such as medical image analysis \cite{kong2024multi, he2023geometric} and autonomous driving \cite{jiang2024senna, cui2024survey}. Prior efforts have sought to mitigate OH in LVLMs through extensive visual instruction fine-tuning \cite{jiang2024hallucination, yu2024hallucidoctor, liu2023mitigating, yuan2024osprey, zhu2025debiasedfinetuningvisionlanguagemodels}, which has proven effective in reducing modality bias \cite{zhu2025debiasedfinetuningvisionlanguagemodels} and enhancing cross-modal alignment \cite{liu2024survey}. However, these approaches typically require substantial computational cost and large-scale training data, severely limiting their scalability and adoption in real-world applications. Other research has aimed to address OH with lower computational overhead by exploiting model attributes and internal states (e.g., model weights, internal tokens, attention maps) \cite{zhuang2025vasparse, wang2025mint, yang2025nullu, an2025mitigating, jiang2025devils, dastmalchi2026fighting, liu2024paying, zhang2026prefilltimeinterventionmitigatinghallucination, lei2026see, ji2026causallens, lin2026hulluedit}.
For instance, VASparse \cite{zhuang2025vasparse} employs token pruning to mitigate OH by reducing reliance on language-biased and low-semantic tokens, and Nullu \cite{yang2025nullu} edits LVLMs via model weight projection to filter out hallucinated features. However, manipulating the internal states or layers of LVLMs often carries the risk of compromising their pre-trained knowledge.  To address this limitation, another line of work \cite{leng2024mitigating, huang2024opera, chen2024halc, favero2024multi}, which is more closely related to ours, has focused on improving the decoding process of LVLMs with minimal or no access to model’s internal states.

\vspace{0.1cm}
 \noindent \textbf{Decoding Validation in LVLMs} aims to mitigate hallucinations at the output stage by leveraging observable signals (e.g., token probabilities or semantic consistency) to flag and correct suspicious objects in generated text. Recently, the methods for improving the decoding process have received great attention. Specifically, OPERA \cite{huang2024opera} investigates token aggregation behaviors, termed “anchor patterns,” that reliably mark the beginning of hallucinated outputs. M3ID \cite{favero2024multi} generates tokens with higher mutual information to the reference image.
Other decoding strategies often use contrastive decoding \cite{chen2024halc, park2025convis, zhu2025ibd, leng2024mitigating, lyu2024alleviating, wu2025season, li2026cross, lei2026see}, contrasting outputs under auxiliary and original inputs as a secondary decoding step to discourage reliance on prior knowledge over visual evidence and thereby suppress ungrounded content.
For instance, VCD \cite{leng2024mitigating} explicitly compares the model’s predictions on the original and noised images to estimate and suppress prior bias, whereas ConVis \cite{park2025convis} and DeGF \cite{zhang2025self} leverage a diffusion model to construct the auxiliary image.
However, these methods often rely on secondary decoding \cite{leng2024mitigating, chen2024halc} or multiple generations from the original prompt \cite{deng2024seeing}, which can limit efficiency. Designing a validator that operates during decoding without extra generations  while maintaining high efficiency represents a promising solution that has not been thoroughly explored. Following this idea, we introduce TTH, a novel test-time object validator. Unlike prior work, TTH explicitly monitors the reliability of each generated token, enabling lightweight validation directly within the description generation process.

\noindent

\begin{figure*}[!t]
	\centering
	\captionsetup{skip=6pt} % Reduce space between image and caption
	\includegraphics[width=1
    \linewidth]{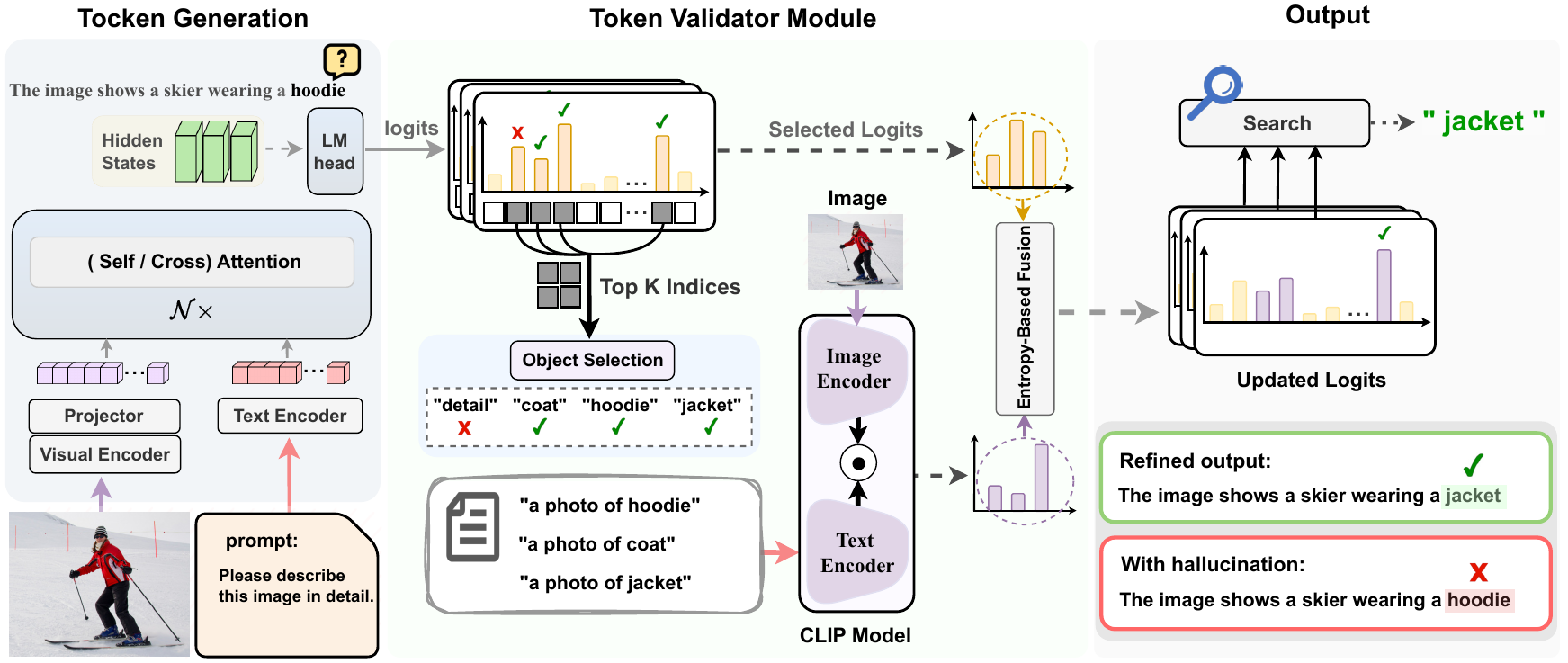}
    \caption{Overview of the proposed Test-Time Hallucination mitigation (TTH) framework. The LVLM first generates token logits conditioned on the image and prompt. The Token Validator Module extracts top-$k$ candidates, filters object tokens using WordNet, and verifies them with CLIP by matching object-specific prompts (e.g., “a photo of jacket”) against the image. Verified scores are rescaled and fused with the LVLM logits using entropy-based weighting. The refined logits yield grounded outputs (e.g., “jacket”) while suppressing hallucinated tokens (e.g., “hoodie”).}
	\label{fig:method}
\end{figure*}

\section{Method}

In this section, we first provide background on LVLMs and the CLIP~\cite{radford2021learning} model as the Zero-Shot Multi-Modal Classifier, followed by the introduction of our proposed method, TTH, for hallucination mitigation in LVLMs.

\subsection{Background}

\noindent{\textbf{Large Vision-Language Models (LVLMs).}} We consider an LVLM parameterized by \(\theta \) which takes an image $\mathbf{I}$ and a query $\mathcal{I}$ as input and transforms them into a sequence of $N$ textual query tokens 
\(
\mathbf{x} = [\mathbf{x}_1, \mathbf{x}_2, \dots, \mathbf{x}_N]
\)
and a sequence of $K$ visual context tokens 
\(
\mathbf{v} = [\mathbf{v}_1, \mathbf{v}_2, \dots, \mathbf{v}_K]
\).
At each decoding step $t$, the LVLM autoregressively generates the last output token:
\begin{equation}
    \mathbf{h}_t = f(\mathbf{x}, \mathbf{v}, \mathbf{y}_{<t}), 
    \quad \mathbf{h}_t \in \mathbb{R}^d,
    \label{eq:1}
\end{equation}
where $\mathbf{y}_{<t} = [\mathbf{y}_1, \dots, \mathbf{y}_{t-1}]$ are the previously generated tokens and $d$ is the hidden dimension.  
The hidden state $\mathbf{h}_t$ is then projected by the language modeling (LM) head into a logit vector:
\begin{equation}
    \mathbf{z}_t = \textbf{W}_{\text{LM}} \mathbf{h}_t, \quad 
    \mathbf{z}_t \in \mathbb{R}^{|\mathcal{V}|},
\end{equation}
where $\textbf{W}_{\text{LM}} \in \mathbb{R}^{|\mathcal{V}| \times d}$ is the LM head projection matrix, and $\mathcal{V}$ denotes the vocabulary.  Finally, the conditional probability of generating token $w \in \mathcal{V}$ at time $t$ is obtained using the softmax function:
\begin{equation}
    p_{\boldsymbol{\theta}}(\mathbf{y}_t = w \mid \mathbf{v}, \mathbf{x}, \mathbf{y}_{<t}) 
    = \frac{\exp(z_{t,w})}{\sum_{w' \in \mathcal{V}} \exp(z_{t,w'})},
    \label{eq:lm_prob}
\end{equation}
where $z_{t,w}$ denotes the logit score corresponding to token $w$.  
The next token $\mathbf{y}_t$ is then sampled (or selected under greedy decoding) from this probability distribution:
\begin{equation}
    \mathbf{y}_t \sim p_{\boldsymbol{\theta}}(\cdot \mid \mathbf{v}, \mathbf{x}, \mathbf{y}_{<t}).
\end{equation}

\vspace{0.1cm}
\noindent{\textbf{Contrastive Language-Image Pretraining (CLIP).}} A complementary approach to visual grounding is provided by zero-shot multi-modal classifiers (MMCs) such as CLIP. 
This model learns a joint embedding space between images and texts through contrastive pretraining. 
CLIP consists of two encoders: a visual encoder $E_v$ and a text encoder $E_t$. 
Given an image $\textbf{I}$ and a candidate textual description $c_i$, the encoders produce embeddings 
$\textbf{u} = E_v(\textbf{I})$ and $\textbf{e}_i = E_t(c_i)$ in a shared space, and the alignment score is computed as 
$\ell_{c_i} =  \text{sim}(\textbf{u}, \textbf{e}_i)$ where \(\text{sim}(.)\) denotes cosine similarity. 
Over a set of $M$ candidate class descriptions  $\{c_1, c_2, \dots, c_M\}$, the scores are normalized into a probability distribution:
\begin{equation}
    p(c_i \mid \textbf{I}) = 
    \frac{\exp(\ell_{c_i}/\tau)}{\sum_{j=1}^M \exp(\ell_{c_j}/\tau)},
\end{equation}
where $\tau$ is a temperature parameter. 
Thus, CLIP enables zero-shot classification by selecting the candidate text most aligned with the given image, 
making it highly effective for grounding object mentions in visual content.

\subsection{Test-Time Hallucination (TTH)}

While LVLMs generate fluent and contextually relevant text, their reliance on language priors often leads to object hallucinations. In contrast, MMCs such as CLIP excel at grounding textual candidates in the visual input but lack the ability to produce free-form responses. To bridge these complementary strengths, we introduce Test-Time Hallucination Mitigation (TTH), a decoding refinement framework that integrates CLIP-based verification into LVLM token generation. At each step, TTH identifies object-related tokens among the top-$k$ candidates, rescoring them with CLIP to ensure object mentions remain visually grounded while preserving the fluency of LVLM outputs. This process is illustrated in Figure \ref{fig:intro}.

\vspace{0.1cm}
\noindent \textbf{Logit Generation and Candidate Selection.} 
At each decoding step $t$, the LVLM produces a logit vector $\textbf{z}_t =[z_{t,w_1},..., z_{t,w_{|\mathcal{V}|}}] \in \mathbb{R}^{|\mathcal{V}|}$ over the vocabulary $\mathcal{V}$ using its LM head. 
These logits correspond to the probability distribution defined in Equation \ref{eq:lm_prob}. 
From $\textbf{z}_t$, we extract the top-$k$ candidate tokens with the highest probabilities:
\begin{equation}
    R_t = \{w_1, w_2, \dots, w_k\} = \text{TopK}(\textbf{z}_t, k).
\end{equation}

\vspace{0.1cm}
\noindent \textbf{Object Candidate Detection.}
We next identify which of the top-$k$ candidates correspond to physical objects. 
This is determined using lexical resources such as WordNet \cite{miller1995wordnet}.
Formally, the object candidate set is
\[
    \mathcal{O}_t = \{ w_i \in R_t \mid w_i \text{ is tagged as an object in WordNet} \}.
\]
If $|\mathcal{O}_t| \leq 1$, no refinement is applied and decoding proceeds normally. 
If multiple object candidates are found, we perform CLIP-based verification.

\vspace{0.1cm}
\noindent \textbf{CLIP-based Object Verification.}
For each object candidate $w_i \in \mathcal{O}_t$, we construct a grounding phrase 
$c_i = \texttt{``a photo of }\left\{ w_i \right\}\texttt{''}$ and encode it with the CLIP text encoder $E_t$, obtaining 
$\textbf{e}_i = E_t(c_i)$.  
The image embedding $\textbf{u} = E_v(\textbf{I})$ is precomputed by the visual encoder $E_v$.  
We then compute the cosine similarity between the image and text embeddings as \(\ell_{w_i} = \text{sim}(\mathbf{u}, \mathbf{e}_i)\). These similarity scores $\ell_{w_i}$ serve as object-level logits that measure how consistent each candidate token is with the visual content.

\vspace{0.1cm}
\noindent \textbf{Logit Rescaling.}  
The original LVLM logit $z_{t,w_i}$ and the CLIP similarity scores $\ell_{w_i}$ are not directly comparable, since they lie on different scales. 
To address this, we rescale $\ell_{w_i}$ into the range of $\{{z}_{t,{w_i}} : w_i \in \mathcal{O}_t\}$:
\begin{equation}
        \tilde{\ell}_{w_i} = \alpha + \frac{(\ell_{w_i} - \min_{w_j \in \mathcal{O}_t} \ell_{w_j})}{(\max_{w_j \in \mathcal{O}_t} \ell_{w_j} - \min_{w_j \in \mathcal{O}_t} \ell_{w_j} + \epsilon)} \, (\beta - \alpha),
\end{equation}
where $\alpha = \min_{w_j \in \mathcal{O}_t} {z}_{t,w_j}$ and $\beta = \max_{w_j \in \mathcal{O}_t} {z}_{t, w_j}$, and $\epsilon$ is a small constant to avoid division by zero. 

\vspace{0.1cm}
\noindent \textbf{Entropy-adaptive Fusion.}  
To balance linguistic priors and visual grounding, we compute the normalized entropy of the LVLM distribution over $\mathcal{O}_t$:
\begin{equation}
\begin{split}
    & p(w_i) = \frac{\exp({z}_{t,w_i})}{\sum_{w_j \in \mathcal{O}_t} \exp({z}_{t,w_j})}, \\
    & H_t= -\frac{1}{\log |\mathcal{O}_t|} \sum_{w_i \in \mathcal{O}_t} p(w_i) \log p(w_i).
\end{split}
\end{equation}
Here $H_t \in [0,1]$ reflects the model’s uncertainty at step $t$. A higher entropy indicates that the LVLM is less confident among object candidates, which correlates with a higher chance of hallucination. In such cases, we assign more weight to CLIP verification. The fusion rule is therefore defined as:
\begin{equation}
    {z}'_{t,w_i} = H_t \cdot \tilde{\ell}_{w_i} + (1 - H_t) \cdot {z}_{t,w_i}.
    \label{eq:9}
\end{equation}

\vspace{0.1cm}
\noindent \textbf{Decoding.}  
The refined logits ${z}'_{t,w_i}$ for $w_i \in \mathcal{O}_t$ replace the original logits of the object candidates, while non-object tokens retain their original values.
These refined logits are then used in the token selection process—optionally employing strategies such as beam search—to generate the final response $\textbf{y}$, resulting in reduced object hallucinations. Algorithm~\ref{alg:1} provides the full description of the method.

\vspace{-0.2cm}

\begin{algorithm}[t]
%  way one:
% \scriptsize

%  way two:
\small 

\caption{Test-Time Hallucination Control}
\label{alg:tth}
% Define Inputs and Outputs
\KwIn{image $\textbf{I}$, query $\mathcal{I}$, LVLM, CLIP visual encoder $E_v$, CLIP text encoder $E_t$, top-$k$ size $k$, and WordNet.}
\KwOut{The generated text $\textbf{y}$}
\BlankLine

% Algorithm body
$\textbf{u} \gets E_v(\textbf{I})$\;
$\textbf{y} \gets [\,]$\;

\For{$t = 1$ to $T$}{
  \textbf{Logit generation:} $\textbf{z}_t \gets  Eq.~(\ref{eq:1})$\;
  \textbf{Candidate extraction:} $R_t \gets \text{TopK}(\textbf{z}_t,k)$\;
  $\mathcal{O}_t \gets \{w_i \in R_t \mid w_i \text{ tagged as object (WordNet)}\}$\;

  \If{$|\mathcal{\mathcal{O}}_t| > 1$}{
    \textbf{CLIP verification:}\;
    \For{$w_i \in \mathcal{O}_t$}{
      $\textbf{e}_i \gets E_t(\texttt{``a photo of }\left\{ w_i \right\}\texttt{''})$\; 
      $\ell_{w_i} \gets \text{sim}(\textbf{u}, \textbf{e}_i)$\;
    }

    \textbf{Rescaling:} normalize $\ell_{w_i}\to \tilde{\ell}_{w_i}$ into range of $\textbf{z}_t[\mathcal{O}_t]$\;

    \textbf{Entropy fusion:}\;
    % $p(w_i)\propto \exp(z_t[w_i])$, \quad
    %        $H_t \gets -\tfrac{1}{\log |O_t|}\sum_{w_i\in O_t} p(w_i)\log p(w_i)$\;
    \For{$w_i \in \mathcal{O}_t$}{
      ${z}'_{t,w_i} \gets H_t \cdot \tilde{\ell}_{w_i}+(1-H_t) \cdot {z}_{t,w_i}$\;
    }
    ${z}'_{t, w_i} \gets {z}_{t, w_i}$ for all $w_i \notin \mathcal{O}_t$\;
  }
  \Else{
    $\textbf{z}'_t \gets \textbf{z}_t$\;
  }

  $\textbf{y}_t \gets \text{SelectNextToken}(\textbf{z}'_t)$; append $\textbf{y}_t$ to $\textbf{y}$\;
  \If{$\textbf{y}_t$ is \textsc{eos}}{
    \textbf{break}\;
  }
}
\label{alg:1}
\KwRet{$\textbf{y}$}\;
\end{algorithm}
\section{Experiments}

We conduct extensive experiments to assess the effectiveness of TTH in mitigating object hallucinations. Our evaluation covers standard benchmarks and includes comparisons with recent state-of-the-art approaches.  Further analysis is presented in Section~\ref{sec:Ablation_study}.

% follosing the \cite{chen2024halc} we used 
% offline POPE (OPOPE)

\subsection{Experimental Setup}

\noindent\textbf{Datasets \& Evaluation Metrics.} We evaluate on the widely adopted CHAIR~\cite{rohrbach2018object}, OPOPE~\cite{chen2024halc}, and LLaVA-Bench~\cite{liu2023improved} benchmarks. For CHAIR and OPOPE, we follow the standard evaluation protocol and report results averaged over three random seeds, each consisting of 500 images sampled from the MSCOCO dataset \cite{lin2014microsoft}. CHAIR measures the alignment between generated captions and ground-truth labels: CHAIR\(_S\) quantifies the proportion of hallucinated sentences, while CHAIR\(_I\) measures the proportion of hallucinated objects across all generated outputs. BLEU is additionally reported to assess overall caption quality. OPOPE evaluates hallucination by analyzing the inclusion of sampled positive and negative objects within captions under three settings: random, popular, and adversarial. LLaVA-Bench serves as a more challenging benchmark, comprising 24 images paired with 60 diverse questions spanning both indoor and outdoor domains. We use all 24 images for our GPT-4-assisted evaluation.

\vspace{0.1cm}
\noindent\textbf{Implementation details.}
The experiments are conducted on three LVLMs, including LLaVA-1.5 \cite{liu2023improved} with Vicuna \cite{vicuna2023}, MiniGPT-4 \cite{zhu2023minigpt4} with Llama2 \cite{touvron2023llama2}, and mPLUG-Owl2 \cite{ye2024mplug}. Following the experimental settings in \cite{yang2025nullu, jiang2025devils, chen2024halc}, we randomly sample images from the COCO 2014 validation set \cite{lin2014microsoft}. For all experiments, the CLIP ViT-B/32 model is employed as the token validator, with the number of object candidates set to 10. To improve efficiency, the image encoder is initialized only once for each image, and the text is processed for all tokens at once for each beam. The number of beams is 3 for LLaVA-1.5 and MiniGPT-4, and 1 for mPLUG-Owl2 as reported in \cite{yang2025nullu}.
All experiments were conducted on a single NVIDIA RTX 4090 GPU.

\vspace{0.1cm}
\noindent\textbf{Baselines.}
To evaluate our TTH and ensure a fair comparison, we compare our method against standard baselines following \cite{yang2025nullu}, including Greedy Search, Beam Search \cite{freitag2017beam}, DoLa \cite{chuang2023dola}, Woodpecker \cite{yin2023woodpecker}, LURE \cite{zhou2023analyzing}, and Nullu \cite{yang2025nullu}. In addition, we assess TTH’s effectiveness in mitigating hallucinations against successful decoding strategies specifically designed to reduce OH, such as VCD \cite{leng2024mitigating}, OPERA \cite{huang2024opera}, and HALC \cite{chen2024halc}.

\subsection{Experimental Results}
\label{subsec:results}

\subsubsection{Results on CHAIR.}
We present experimental results on the CHAIR benchmark in Table~\ref{tab:results-chair}, which show that TTH consistently outperforms baseline methods across all source LVLMs. For both LLaVA-1.5 and MiniGPT-4, the improvements are particularly pronounced on CHAIR\(_I\) and CHAIR\(_S\), indicating a substantial reduction in hallucinated objects while maintaining the BLEU score. Compared to Nullu~\cite{yang2025nullu}, TTH achieves higher performance across all models without requiring access to model weights; instead, it operates on-the-fly during decoding at inference time. Furthermore, TTH surpasses widely used decoding-based approaches such as OPERA~\cite{leng2024mitigating}, HALC~\cite{chen2024halc}, and VCD~\cite{huang2024opera}, delivering a clear margin of improvement. Consistent gains on MiniGPT-4 and mPLUG-Owl2 further confirm that TTH effectively suppresses OH in off-the-shelf LVLMs while preserving fluency and overall generation quality.

\begin{table*}[!t]
\centering
\captionsetup{skip=6pt} % 
\resizebox{\textwidth}{!}{%
\setlength{\tabcolsep}{4pt}
\renewcommand{\arraystretch}{1.3}
\begin{tabular}{l|ccc|ccc|ccc}
   \Xhline{3\arrayrulewidth}
\textbf{Method} & \multicolumn{3}{c|}{\textbf{LLaVA-1.5}} & \multicolumn{3}{c|}{\textbf{MiniGPT-4}} & \multicolumn{3}{c}{\textbf{mPLUG-Owl2}} \\
& \textbf{Accuracy$\uparrow$} & \textbf{Precision$\uparrow$} & \textbf{F score$\uparrow$} & \textbf{Accuracy$\uparrow$} & \textbf{Precision$\uparrow$} & \textbf{F score$\uparrow$} & \textbf{Accuracy$\uparrow$} & \textbf{Precision$\uparrow$} & \textbf{F score$\uparrow$} \\
\Xhline{2\arrayrulewidth}
\textbf{Greedy} & 79.14{\tiny$\pm$0.89} & 91.98{\tiny$\pm$0.82} & 90.45{\tiny$\pm$0.86} & 71.22{\tiny$\pm$1.27} & 93.72{\tiny$\pm$1.02} & 90.04{\tiny$\pm$1.23} & 76.46{\tiny$\pm$0.92} & 88.85{\tiny$\pm$1.15} & 87.29{\tiny$\pm$1.15} \\
\textbf{Beam Search} \cite{freitag2017beam} & 79.41{\tiny$\pm$0.69} & 92.52{\tiny$\pm$0.55} & 90.96{\tiny$\pm$0.59} & 71.65{\tiny$\pm$1.15} & 94.70{\tiny$\pm$0.60} & 90.97{\tiny$\pm$0.85} & 76.76{\tiny$\pm$1.02} & 90.28{\tiny$\pm$0.80} & 88.56{\tiny$\pm$0.87} \\
\textbf{DoLa} \cite{chuang2023dola} \small{(ICLR'24)} & 78.98{\tiny$\pm$0.56} & 91.66{\tiny$\pm$0.81} & 90.15{\tiny$\pm$0.79} & 71.28{\tiny$\pm$1.15} & 93.92{\tiny$\pm$0.83} & 90.22{\tiny$\pm$1.04} & 76.07{\tiny$\pm$1.09} & 88.54{\tiny$\pm$1.25} & 86.95{\tiny$\pm$1.27} \\
\textbf{OPERA} \cite{huang2024opera}  \small{(CVPR'24)} & 79.29{\tiny$\pm$0.32} & 92.25{\tiny$\pm$0.07} & 90.71{\tiny$\pm$0.11} & 70.48{\tiny$\pm$1.63} & 94.41{\tiny$\pm$1.11} & 90.66{\tiny$\pm$1.42} & 75.49{\tiny$\pm$1.29} & 91.23{\tiny$\pm$1.06} & 89.11{\tiny$\pm$1.17} \\
\textbf{VCD} \cite{leng2024mitigating} \small{(CVPR'24)} & 78.01{\tiny$\pm$0.75} & 91.33{\tiny$\pm$0.88} & 89.69{\tiny$\pm$0.89} & 70.83{\tiny$\pm$1.83} & 92.31{\tiny$\pm$0.88} & 88.76{\tiny$\pm$1.29} & 75.49{\tiny$\pm$1.27} & 88.75{\tiny$\pm$1.56} & 87.02{\tiny$\pm$1.57} \\
\textbf{HALC} \cite{chen2024halc} \small{(ICML'24)} & 77.87{\tiny$\pm$0.22} & 93.17{\tiny$\pm$0.39} & 91.25{\tiny$\pm$0.38} & 71.17{\tiny$\pm$0.89} & 94.88{\tiny$\pm$0.15} & 90.95{\tiny$\pm$0.42} & 74.93{\tiny$\pm$1.09} & 90.20{\tiny$\pm$0.90} & 88.12{\tiny$\pm$0.99} \\

\textbf{Nullu} \cite{yang2025nullu}\small{(CVPR'25)}
       & 79.52{\tiny$\pm$0.04} & 93.46{\tiny$\pm$0.03} & 91.79{\tiny$\pm$0.04} 
       & 71.92{\tiny$\pm$0.39} & 95.96{\tiny$\pm$0.65} & 92.07{\tiny$\pm$0.65} 
       & 77.09{\tiny$\pm$1.37} & 92.83{\tiny$\pm$0.29} & 90.80{\tiny$\pm$0.52} \\

\midrule
\ccol \textbf{TTH (Ours)} & \ccol \textbf{80.46}{\tiny$\pm 0.04 $}& \ccol \textbf{93.67}{\tiny$\pm$0.03} & \ccol \textbf{92.23}{\tiny$\pm$0.06} & \ccol \textbf{72.6}{\tiny$\pm$0.31} & \ccol \textbf{96.23}{\tiny$\pm$0.18} & \ccol \textbf{92.81}{\tiny$\pm$ 0.35} & \ccol \textbf{77.89}{\tiny$\pm$0.45} & \ccol \textbf{92.97}{\tiny$\pm$0.21} & \ccol \textbf{91.40}{\tiny$\pm$0.52} \\
         \bottomrule
    \Xhline{3\arrayrulewidth}
\end{tabular}%
}
\caption{Accuracy, precision, and F1 scores of different methods on the OPOPE benchmark.}
\label{tab:opope}
\end{table*}

\captionsetup{skip=6pt} 
\begin{table*}[!t]
\centering
\renewcommand{\arraystretch}{1.25}
\resizebox{\textwidth}{!}{%
\begin{tabular}{l|ccc|ccc|ccc}
    \Xhline{3\arrayrulewidth}

\textbf{Method} & \multicolumn{3}{c|}{\textbf{LLaVA-1.5}} & \multicolumn{3}{c|}{\textbf{MiniGPT-4}} & \multicolumn{3}{c}{\textbf{mPLUG-Owl2}} \\
 & \textbf{CHAIR$_S \downarrow$} & \textbf{CHAIR$_I \downarrow$} & \textbf{BLEU$\uparrow$}
 & \textbf{CHAIR$_S \downarrow$} & \textbf{CHAIR$_I \downarrow$} & \textbf{BLEU$\uparrow$}
 & \textbf{CHAIR$_S \downarrow$} & \textbf{CHAIR$_I \downarrow$} & \textbf{BLEU$\uparrow$} \\
\Xhline{2\arrayrulewidth}
\textbf{Greedy} & 20.40$_{\pm2.80}$ & 7.08$_{\pm0.33}$ & 15.72$_{\pm0.10}$ 
       & 32.40$_{\pm2.20}$ & 12.20$_{\pm0.42}$ & 14.57$_{\pm0.11}$ 
       & 22.90$_{\pm0.90}$ & 8.62$_{\pm0.11}$ & 15.01$_{\pm0.24}$ \\
\textbf{Beam Search } \cite{freitag2017beam}
       & 19.50$_{\pm2.30}$ & 6.84$_{\pm0.79}$ & 15.99$_{\pm0.14}$ 
       & 30.10$_{\pm0.30}$ & 11.87$_{\pm0.37}$ & 15.35$_{\pm0.24}$ 
       & 20.30$_{\pm0.70}$ & 7.62$_{\pm0.19}$ & 15.43$_{\pm0.05}$ \\
\textbf{DoLa} \cite{chuang2023dola} \small{(ICLR'24)}
       & 20.20$_{\pm2.80}$ & 6.75$_{\pm0.54}$ & 15.68$_{\pm0.10}$ 
       & 31.90$_{\pm3.30}$ & 12.15$_{\pm0.89}$ & 14.54$_{\pm0.12}$ 
       & 22.40$_{\pm1.80}$ & 8.36$_{\pm0.04}$ & 15.13$_{\pm0.21}$ \\
\textbf{OPERA} \cite{huang2024opera} \small{(CVPR'24)}
       & 17.50$_{\pm0.50}$ & 6.07$_{\pm0.32}$ & 16.02$_{\pm0.02}$ 
       & 29.70$_{\pm0.30}$ & 11.96$_{\pm0.29}$ & 14.82$_{\pm0.05}$ 
       & 20.07$_{\pm2.07}$ & 7.18$_{\pm0.39}$ & 15.41$_{\pm0.12}$ \\
\textbf{VCD} \cite{leng2024mitigating} \small{(CVPR'24)}
       & 20.30$_{\pm1.10}$ & 7.28$_{\pm0.10}$ & 14.53$_{\pm0.01}$ 
       & 29.00$_{\pm2.80}$ & 12.64$_{\pm1.19}$ & 14.42$_{\pm0.01}$ 
       & 22.80$_{\pm0.80}$ & 8.68$_{\pm0.17}$ & 15.14$_{\pm0.13}$ \\
\textbf{Woodpecker} \cite{yin2023woodpecker} \small{(SCIS'24)}
       & 23.85$_{\pm4.62}$ & 7.50$_{\pm0.01}$ & 17.05$_{\pm0.00}$ 
       & 28.87$_{\pm2.20}$ & 10.20$_{\pm0.85}$ & 15.30$_{\pm0.01}$ 
       & 26.33$_{\pm1.98}$ & 8.43$_{\pm0.80}$ & 16.43$_{\pm0.00}$ \\
\textbf{LURE} \cite{zhou2023analyzing} \small{(ICLR'24)}
       & 19.48$_{\pm2.35}$ & 6.50$_{\pm0.38}$ & 15.97$_{\pm0.01}$ 
       & 27.88$_{\pm2.25}$ & 10.20$_{\pm0.85}$ & 15.03$_{\pm0.01}$ 
       & 21.27$_{\pm0.06}$ & 7.67$_{\pm0.16}$ & 15.65$_{\pm0.15}$ \\
\textbf{HALC} \cite{chen2024halc}\small{(ICML'24)}
       & 16.90$_{\pm2.10}$ & 5.72$_{\pm0.55}$ & 16.02$_{\pm0.04}$ 
       & 25.20$_{\pm2.00}$ & 9.42$_{\pm0.41}$ & 14.91$_{\pm0.13}$ 
       & 18.80$_{\pm1.20}$ & 7.00$_{\pm0.01}$ & 15.33$_{\pm0.24}$ \\

\textbf{Nullu} \cite{yang2025nullu}\small{(CVPR'25)}
       & 15.20$_{\pm0.60}$ & 5.30$_{\pm0.03}$ & 15.69$_{\pm0.04}$ 
       & 21.40$_{\pm1.00}$ & 8.99$_{\pm0.36}$ & 14.81$_{\pm0.06}$ 
       & 15.60$_{\pm1.20}$ & 5.77$_{\pm0.01}$ & 15.45$_{\pm0.01}$ \\

\midrule
\ccol \textbf{TTH (Ours)} & \ccol \textbf{14.27}$_{\pm 0.46}$  & \ccol \textbf{4.76}$_{\pm 0.02}$ & \ccol 15.97 $_{\pm 0.02}$ 
              & \ccol \textbf{20.25}$_{\pm 0.86}$ & \ccol \textbf{8.16}$_{\pm 0.22}$ & \ccol {15.77}$_{\pm 0.19}$
              & \ccol \textbf{15.24}$_{\pm 0.32}$ & \ccol \textbf{5.43}$_{\pm  0.02}$ & \ccol 15.76$_{\pm 0.15}$ \\
         \bottomrule
    \Xhline{3\arrayrulewidth}
    \end{tabular}
}
\vspace{4.5pt}
\caption{CHAIR and BLEU scores across LVLMs; lower CHAIR = less hallucination, higher BLEU = better fluency.}
\label{tab:results-chair}
\end{table*}

\subsubsection{Results on OPOPE.}  The average OPOPE results for three policies (random, adversarial, and popular) are summarized in Table \ref{tab:opope}. Since OPOPE is a particularly challenging metric that requires a holistic image description followed by the identification of a specific object, which is often omitted, its evaluation serves as a rigorous test of model robustness. According to the results in Table \ref{tab:opope}, our observations further indicate that employing a robust token validator can refine suspicious objects during generation and correct them when necessary. Across all settings, our proposed TTH consistently outperforms existing methods in terms of accuracy, precision, and F-score, demonstrating its effectiveness in mitigating OH and its broad applicability across different LVLMs. For instance, as shown in Table \ref{tab:opope}, while other decoding-based baselines \cite{leng2024mitigating, chen2024halc, huang2024opera} struggle to improve OPOPE accuracy when mitigating OH, our TTH successfully enhances the performance of source models: LLaVA-1.5 by 1.32\%, MiniGPT-4 by 1.38\%, and mPLUG-Owl2 by 1.43\%, outperforming all baselines. Similar trends are observed in the Precision and F-score metrics. Consequently,
this increase in F-score further highlights TTH’s enhanced ability to accurately detect object presences.

\subsubsection{GPT-4V Aided Evaluation on LLaVA-Bench.}
\begin{figure*}[!t]
    \centering
\captionsetup{skip=6pt}      \includegraphics[width=\linewidth]{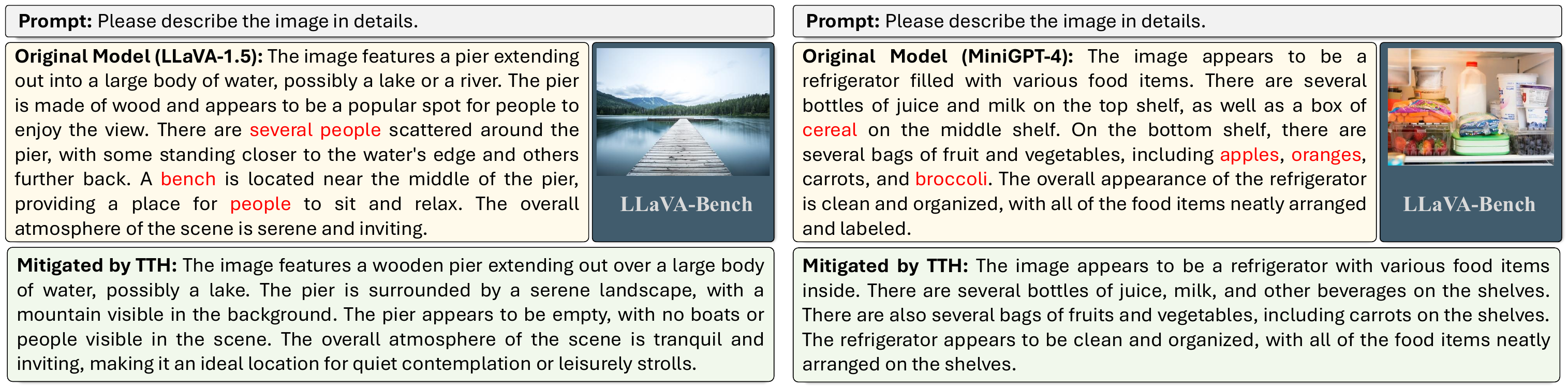}
    \caption{Illustration on LLaVA-Bench: TTH curbs hallucinations and improves grounding quality.}
    \label{fig:llava-bench-example}
\end{figure*}

To rigorously assess the impact of our hallucination mitigation method, we evaluate performance on LLaVA-Bench \cite{liu2023improved} using both quantitative metrics and qualitative examples. For the quantitative study, GPT-4V is prompted to score each response on a 1–10 scale along the dimensions of accuracy and level of detail, conditioned on the given image, question, and model output. As summarized in Table~\ref{tab:llava-bench}, integrating TTH consistently yields higher scores across all three evaluated models, indicating stronger visual grounding and fewer hallucinations. Complementing this, Figure~\ref{fig:llava-bench-example} presents two qualitative examples: the left panel shows results from LLaVA-1.5-7B, while the right panel corresponds to MiniGPT-4. In the LLaVA-1.5 baseline, the generated caption includes hallucinated elements such as “people” and “bench”, whereas the TTH-refined output provides an accurate description consistent with the image. Likewise, the MiniGPT-4 baseline introduces multiple spurious objects, including “cereal", “apples", “oranges", and “broccoli", which are effectively suppressed by the TTH-enhanced model.

\begin{table}[!t]
\centering
\captionsetup{skip=6pt} % 
\small
\renewcommand{\arraystretch}{1.1}
\setlength{\tabcolsep}{7.5pt}
\begin{tabular}{llcc}
\toprule
\textbf{Model} & \textbf{Method} & \textbf{Accuracy$\uparrow$} & \textbf{Detailedness$\uparrow$} \\
\midrule
\multirow{2}{*}{LLaVA-1.5} & Original &  6.46 & 6.25 \\
                           & \ccol \textbf{TTH} & \ccol \ccol \textbf{6.71} & \ccol \textbf{6.46} \\
\midrule
\multirow{2}{*}{MiniGPT-4} & Original & 5.70 & 6.04\\
                           & \ccol \textbf{TTH} & \ccol \textbf{5.96} & \ccol \textbf{6.30} \\
\midrule
\multirow{2}{*}{mPLUG-Owl2} & Original & 5.62 & 5.54 \\
                            & \ccol \textbf{TTH} & \ccol \textbf{6.12} & \ccol \textbf{6.08} \\
\bottomrule
\end{tabular}

\caption{LLaVA-Bench results with GPT-4V evaluations. TTH consistently improves accuracy and detailedness across models.}
\label{tab:llava-bench}
\end{table}

\subsubsection{{Inference Efficiency.}}
\begin{figure}[t!]
	\centering
	\captionsetup{skip=6pt} % Reduce space between image and caption
	% \includegraphics[height=2.in,width=1
 %    \linewidth]{figures/chairs_vs_alpha_symlog.pdf}, trim={16pt 10pt 10pt  1pt}, clip
    
    \includegraphics[width=0.68\linewidth]{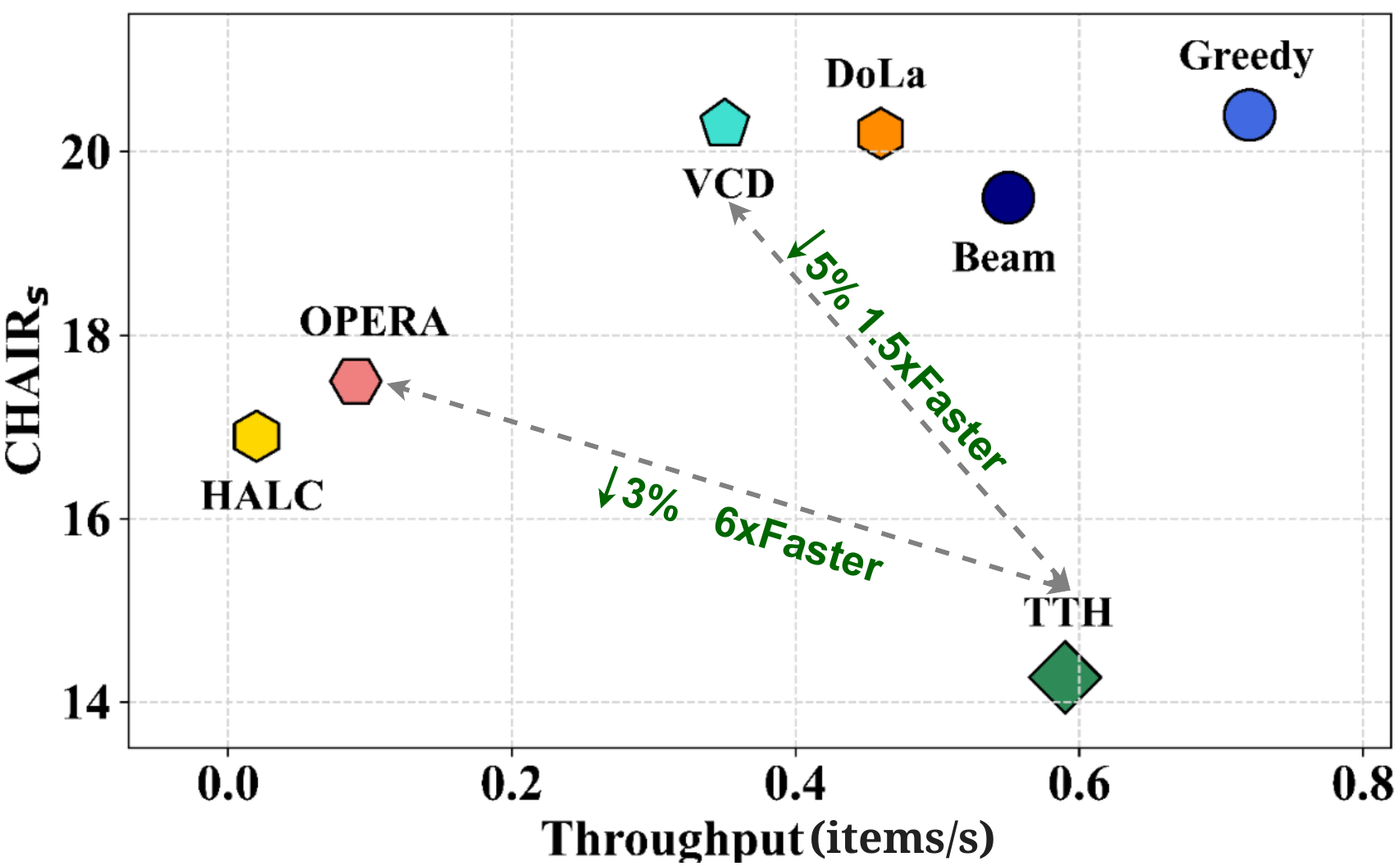}
    % \hfill
    % \includegraphics[width=1\linewidth]{figures/chairi_vs_objects_symlog (cropped) (pdfresizer.com) (1).pdf}
    \caption{Throughput (\(\text{items}/s\)) vs. CHAIR\(_S\) comparison over decoding-based baseline methods. Greedy search is used for TTH in the throughput evaluation. Higher throughput indicates better efficiency, while lower CHAIR\(_S\) reflects reduced hallucination.}

    \label{fig:Throughput}
\end{figure}

To evaluate the efficiency of TTH, we measure inference throughput (\(\text{items}/s\)) alongside the CHAIR\(_S\) metric on LLaVA-1.5-7B for all decoding-based baselines. Throughput is defined as the number of test items processed per second, where higher values indicate faster inference. Results are presented in Figure~\ref{fig:Throughput}.  
Compared to zero-shot greedy decoding (\(0.72\ \text{items}/s\)), TTH achieves a competitive throughput of \(0.59\ \text{items}/s\), incurring only a modest slowdown due to the additional candidate selection and inference latency introduced by the MMC. In contrast, methods such as OPERA (\(0.09\ \text{items}/s\)) and HALC (\(0.02\ \text{items}/s\)) suffer from severe efficiency bottlenecks despite their hallucination reduction.  
Importantly, TTH achieves the lowest CHAIR\(_S\) score (\(14.27\)), demonstrating superior hallucination mitigation while maintaining nearly the same level of efficiency as the greedy baseline. These results highlight TTH as a practical approach that balances strong grounding improvements with minimal computational overhead, outperforming prior methods in both hallucination reduction and throughput trade-offs.

\vspace{0.1cm}

\begin{figure}[t!]
	\centering
	\captionsetup{skip=5pt} % Reduce space between image and caption
	
	\includegraphics[width=0.49\linewidth]{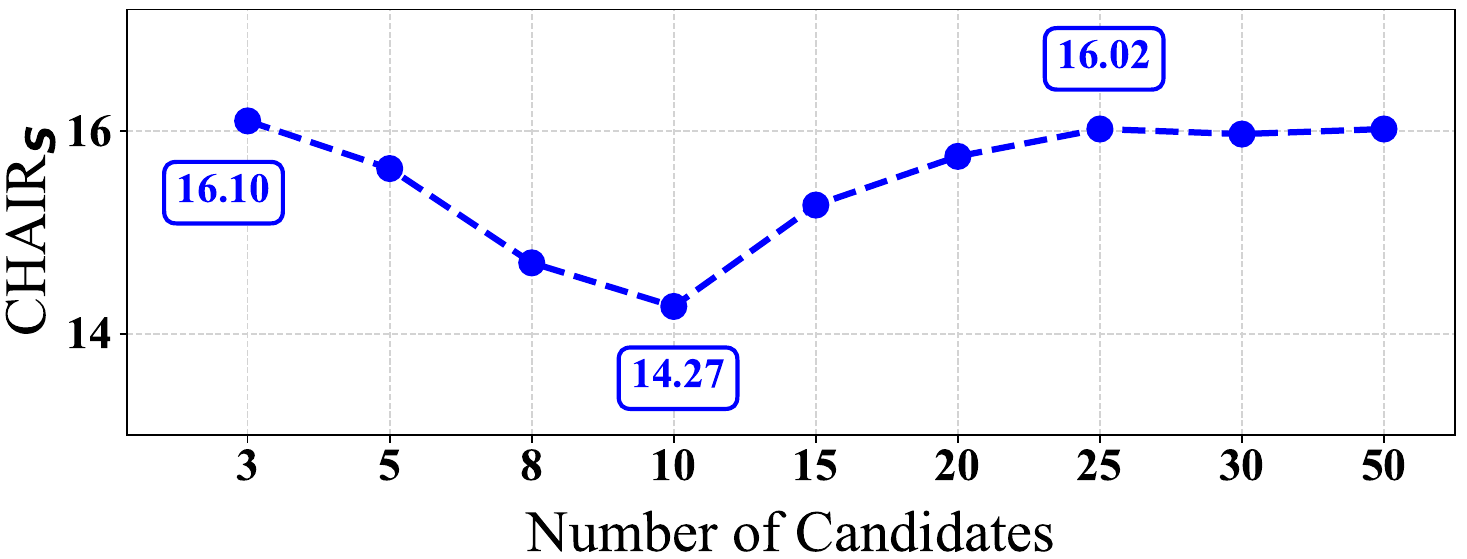}%
	\hfill%
	\includegraphics[width=0.49\linewidth, trim={3pt 0pt 3pt 0pt}, clip]{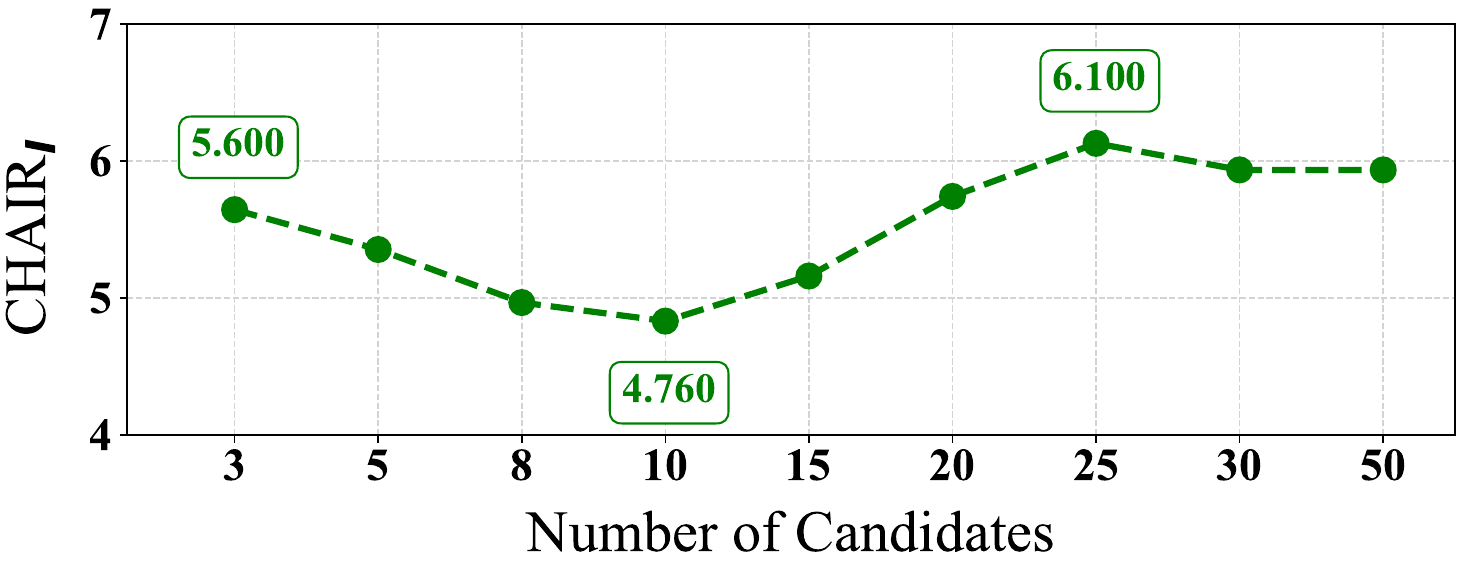}
	
	\caption{Ablation study on the number of object candidates using LLaVA-1.5. Left: CHAIR\(_S\); Right: CHAIR\(_I\).} 
	\label{fig:ablation_object_candidate}
\end{figure}

\subsection{Ablation Study}
\label{sec:Ablation_study}

\textbf{Number of Candidates.}
To investigate the impact of the number of object candidates on hallucination mitigation, we evaluate our method against the CHAIR\(_I\) and CHAIR\(_S\) metrics. As illustrated in Figure~\ref{fig:ablation_object_candidate}, the CHAIR\(_S\) metric decreases as the candidate pool is relatively small, reaching its lowest value at 10 candidates (0.1427). The {CHAIR\(_I\)} metric follows a similar trend, also attaining its minimum at 10 candidates (0.0476). Using too few candidates (e.g., 3) limits the token validator’s effectiveness, as feedback is applied based on a small set of candidates and the correct object may not be present. Conversely, using too many candidates (e.g., 50) degrades performance (CHAIR\(_S\)=0.1602), as it can mislead the MMCs, introduce bias toward dominant objects, and reduce candidate reliability due to top-\(k\) selection. Considering a very wide candidate pool may also increase classifier errors or reinforce biases toward unreliable dominant objects. Overall, setting the number of object candidates to around 10 achieves a balance between candidate diversity and reliability, providing optimal mitigation performance.

\begin{figure}[t!]
	\centering
	\captionsetup{skip=6pt}
	
	% This minipage acts as a "container" for both images.
	% Change 0.8\linewidth to 0.7 or 0.9 to make the whole set smaller or larger.
	\begin{minipage}{1\linewidth} 
		\centering
		\includegraphics[width=0.48\linewidth]{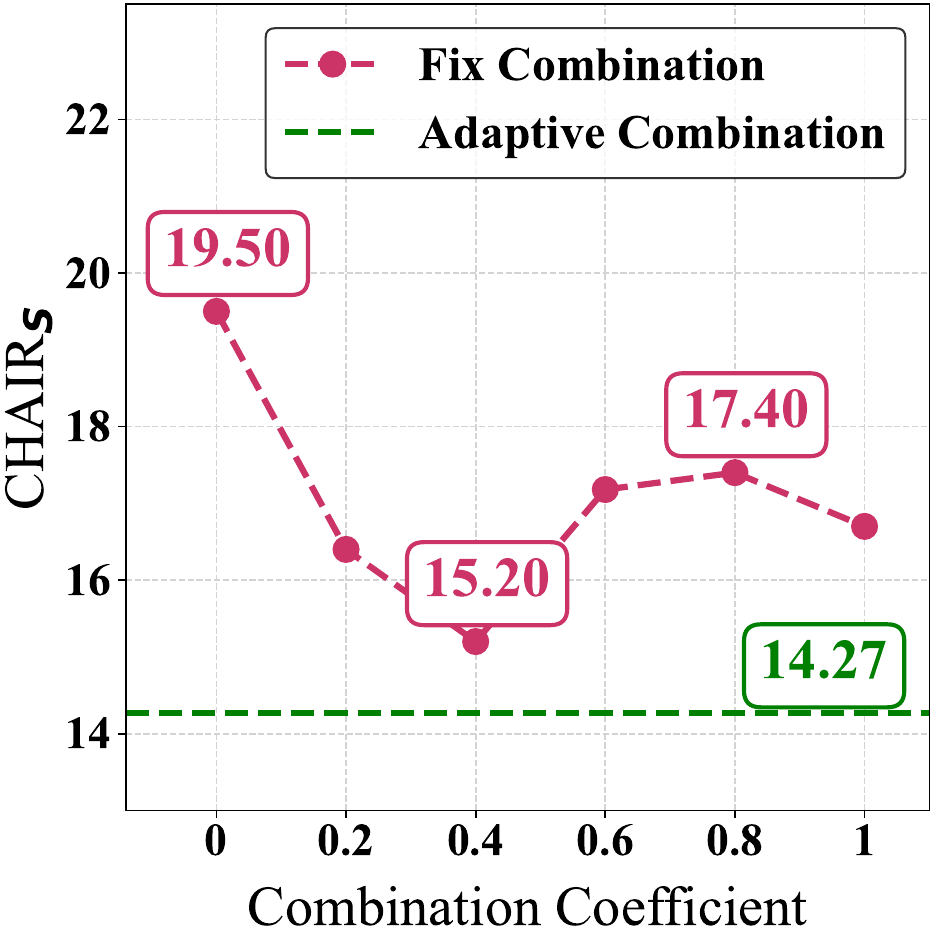}%
		\hfill%
		\includegraphics[width=0.48\linewidth]{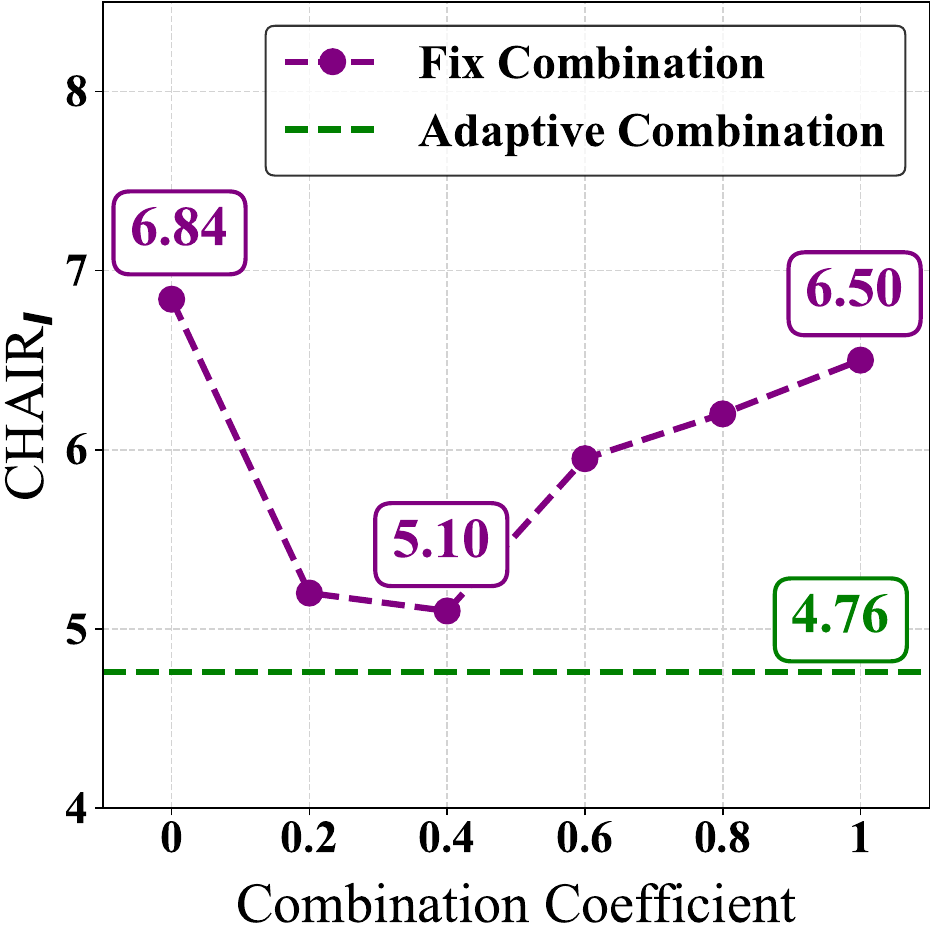}
	\end{minipage}
	
	\caption{Ablation study on the effectiveness of Adaptive Entropy-Based Fusion in LLaVA-1.5. Left: CHAIR\(_S\); Right: CHAIR\(_I\).}
	\label{fig:ablation_entropy_based}
\end{figure}

\vspace{0.1cm}
\noindent\textbf{Entropy-based vs Fixed Fusion.} To examine the effect of using fixed versus adaptive coefficients in the convex combination of the original LM head logits and the validation logits from Equation~\ref{eq:9}, we conducted an ablation study on LLaVA-1.5. Figure~\ref{fig:ablation_entropy_based} shows the results of this experiment. In this setup, the horizontal axis corresponds to the contribution coefficient of the MMC’s logits: when the coefficient is set to zero, the predictions reduce to the original LVLM outputs without adaptation, whereas a coefficient of one implies that the MMC fully specifies the candidate logits, overriding the LVLM’s prior knowledge. The latter often degrades performance, as the classifier may bias toward incorrect classes.  

The experimental results show that fixed coefficients provide some improvements when the original and validation logits contribute more equally. However, they remain sensitive to the choice of coefficient, leading to suboptimal results across settings. By contrast, the entropy-based adaptive fusion strategy     \cite{tamjidi2026adapt} yields consistently better performance (green dashed line), achieving lower CHAIR\(_S\) and CHAIR\(_I\) scores. This adaptive mechanism dynamically increases the influence of validation feedback when the LVLM exhibits higher uncertainty in candidate selection---a scenario closely linked to object hallucination~\cite{jiang2024interpreting}.  
Therefore, TTH adopts this adaptive entropy-based fusion strategy to effectively balance LVLM prior knowledge with MMC validation, providing a more reliable and robust defense against hallucination during decoding.

\section{Limitations and Future Work}

Despite its demonstrated advantages, TTH is primarily tailored to mitigating object hallucinations in image captioning tasks. TTH’s effectiveness is limited in tasks without explicit object generation (e.g., binary probing), as it lacks object-level validation. Additionally, it does not yet address attribute hallucinations, such as incorrect colors or sizes. For future work, we plan to extend TTH beyond object hallucinations by incorporating attribute-level validation. This could be achieved by first detecting candidate attributes and pairing them with corresponding objects to form an object–attribute pool, which can then be passed through the MMC module for joint validation. This extension would allow TTH to address both object- and attribute-level hallucinations, expanding its scope across vision–language tasks. 

\section{Conclusion}

In this paper, we introduced Test-Time Hallucination (TTH), a lightweight, plug-and-play decoding strategy that mitigates object hallucinations in large vision–language models. TTH validates candidate object tokens during decoding using a zero-shot multi-modal classifier and fuses the resulting feedback logits with the original logits via entropy-adaptive weighting. This effectively reduces hallucinations across diverse LVLMs while maintaining efficiency, incurring only minimal computational overhead, and requiring no fine-tuning or modification of model parameters. Extensive experiments demonstrate that TTH consistently improves the reliability and factual accuracy of LVLM outputs. We believe this direction offers a practical and robust solution for real-world vision–language applications.

\clearpage

\bibliographystyle{splncs04}
\bibliography{main}
\end{document}